\documentclass[journal,twoside,web]{ieeecolor}

\usepackage{generic}
\usepackage{cite}
\usepackage{amsmath,amssymb,amsfonts}
\usepackage{algorithmic}
\usepackage{graphicx}
\usepackage{algorithm,algorithmic}
\usepackage{hyperref}
\hypersetup{hidelinks=true}
\usepackage{textcomp}
\usepackage{booktabs}
\usepackage{multirow}
\usepackage{graphicx}
\usepackage{etoolbox}
\usepackage{subfigure}
\usepackage{subfloat}
\def\BibTeX{{\rm B\kern-.05em{\sc i\kern-.025em b}\kern-.08em
    T\kern-.1667em\lower.7ex\hbox{E}\kern-.125emX}}
\begin{document}
\title{TOGS: Gaussian Splatting with Temporal Opacity Offset for Real-Time 4D DSA Rendering}
\author{Shuai Zhang, Huangxuan Zhao, Zhenghong Zhou, Guanjun Wu, Chuansheng Zheng, Xinggang Wang, \IEEEmembership{Member, IEEE}, and Wenyu Liu, \IEEEmembership{Senior Member, IEEE}
\thanks{This work was supported by National Natural Science Foundation of China (No. 62376102). (Corresponding author: Wenyu Liu.)}
\thanks{Shuai Zhang, Zhenghong Zhou, Xinggang Wang, and Wenyu Liu are with the School of Electronic Information and Communications, Huazhong University of Science and Technology, Wuhan, 430074, China. (e-mail: shuaizhang@hust.edu.cn; zhouzhenghong1999@gmail. com; xgwang@hust.edu.cn; liuwy@hust.edu.cn). }
\thanks{Guanjun Wu is with the School of Computer Science \&Technology, Huazhong University of Science and Technology, Wuhan, 430074, China. (e-mail: guajuwu@hust.edu.cn). }
\thanks{Huangxuan Zhao and Chuansheng Zheng are with the Department of Radiology, Union Hospital, Tongji Medical College, Huazhong University of Science and Technology, Wuhan, 430022, China. (e-mail: zhao\_huangxuan@sina.com; hqzcsxh@sina.com).}}

\maketitle

\begin{abstract}
Four-dimensional Digital Subtraction Angiography (4D DSA) is a medical imaging technique that provides a series of 2D images captured at different stages and angles during the process of contrast agent filling blood vessels. It plays a significant role in the diagnosis of cerebrovascular diseases. Improving the rendering quality and speed under sparse sampling is important for observing the status and location of lesions. The current methods exhibit inadequate rendering quality in sparse views and suffer from slow rendering speed. To overcome these limitations, we propose TOGS, a Gaussian splatting method with opacity offset over time, which can effectively improve the rendering quality and speed of 4D DSA. We introduce an opacity offset table for each Gaussian to model the opacity offsets of the Gaussian, using these opacity-varying Gaussians to model the temporal variations in the radiance of the contrast agent. By interpolating the opacity offset table, the opacity variation of the Gaussian at different time points can be determined. This enables us to render the 2D DSA image at that specific moment. Additionally, we introduced a Smooth loss term in the loss function to mitigate overfitting issues that may arise in the model when dealing with sparse view scenarios. During the training phase, we randomly prune Gaussians, thereby reducing the storage overhead of the model. The experimental results demonstrate that compared to previous methods, this model achieves state-of-the-art render quality under the same number of training views. Additionally, it enables real-time rendering while maintaining low storage overhead. The code is available at https://github.com/hustvl/TOGS.
\end{abstract}


\section{Introduction}
\label{sec:introduction}
\IEEEPARstart{T}{he} 4D-DSA volume can not only be viewed at any desired angle but also at any desired time during the passage of the contrast bolus through the vasculature~\cite{ruedinger20214d}. It has been used in the diagnosis of some vascular abnormalities in clinics, such as aneurysms and AVMs/AVFs~\cite{lang20174d}. Compared to 2D and 3D DSA, 4D DSA incorporates temporal information that enables visualization of various abnormal features, such as the flow sequence and velocity of blood entering and leaving abnormal regions. This is crucial for the diagnosis of certain cerebrovascular diseases.

\begin{figure}[!t]
\centering
\hspace{0.42cm}
\subfigure{\includegraphics[width=3.3in]{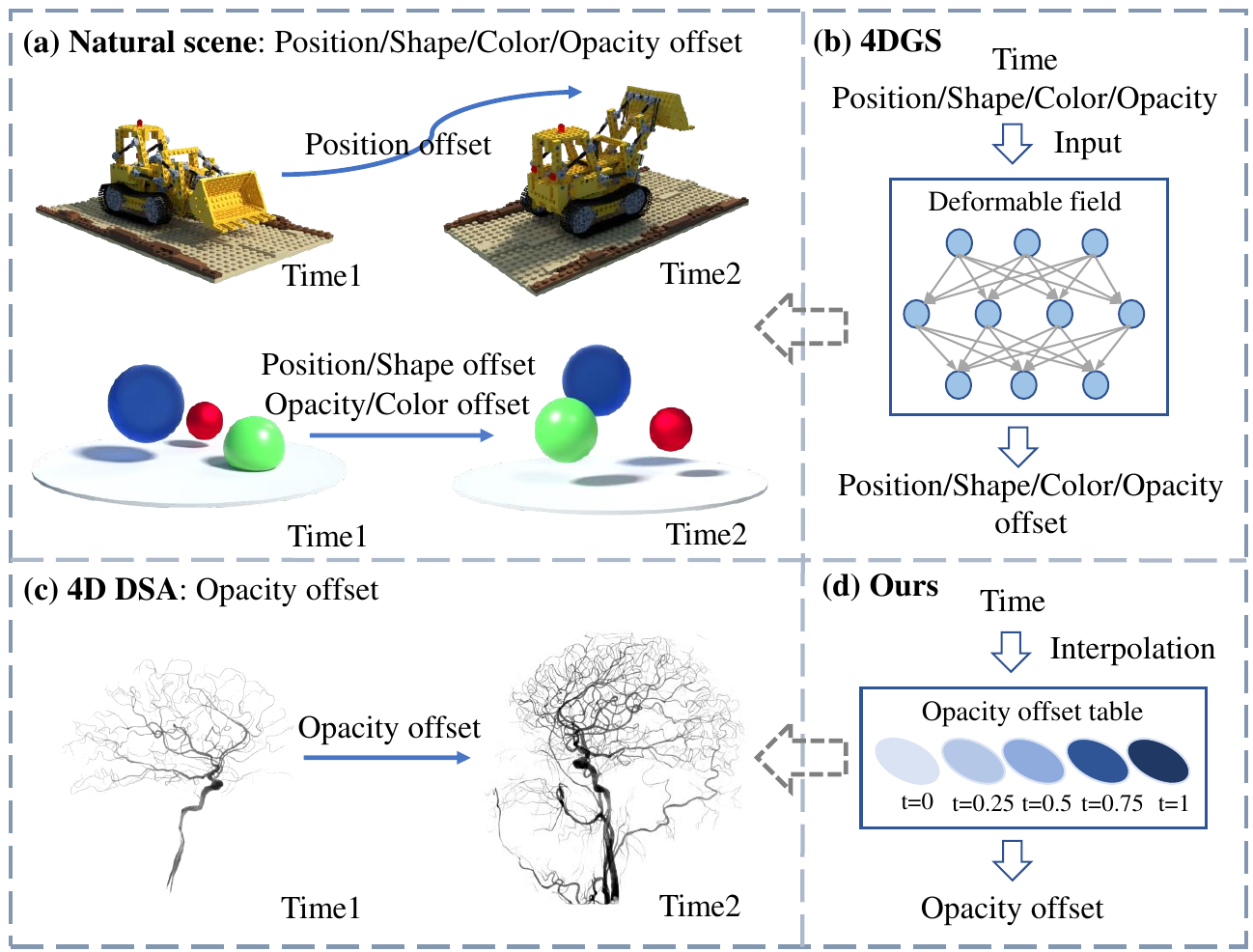}}
\caption{Comparison of temporal variations between natural scenes and 4D DSA scenes.}
\label{fig1}
%\vspace{-10pt}
\hspace{-0.1cm}
\subfigure{\includegraphics[width=3.5in]{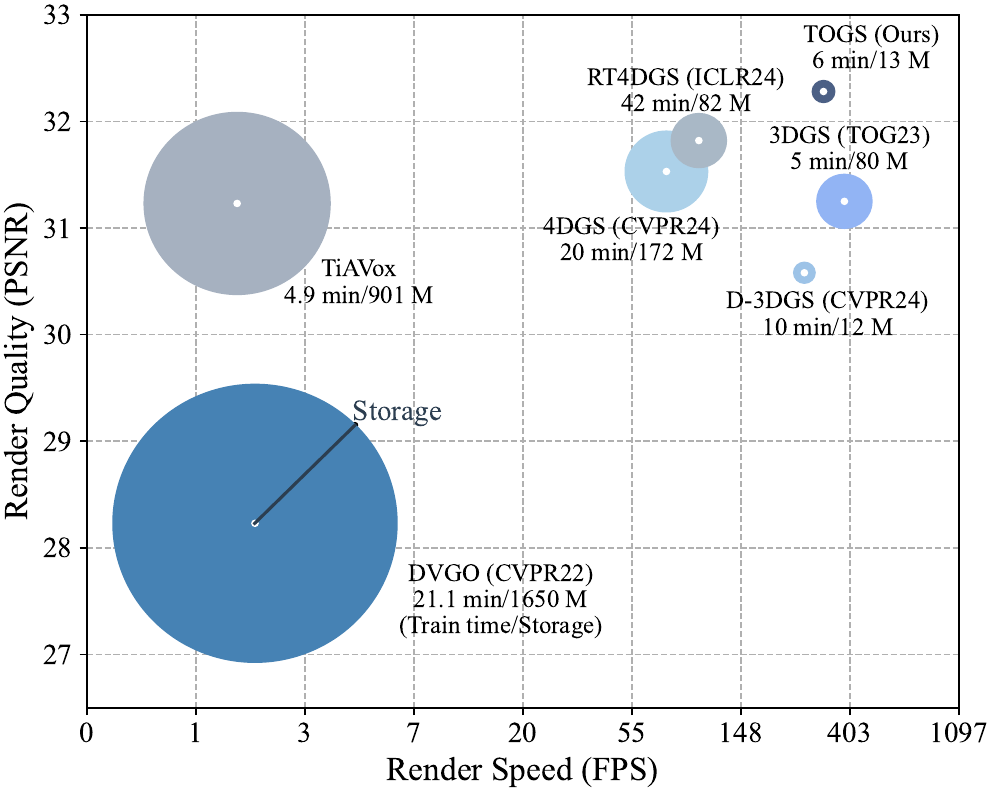}}
\vspace{-15pt}
\caption{Performance of different methods in 4D DSA novel view synthesis (30 training views).}
\label{fig2}
\vspace{-15pt}
\end{figure}

\iffalse
\begin{figure}[!t]
\centering
\hspace{0.42cm}
\includegraphics[width=3.4in]{fig1.pdf}
\caption{Comparison of temporal variations between natural scenes and 4D DSA scenes}
\label{fig1}
%\vspace{-10pt}
\vspace{-15pt}
\end{figure}
\fi
Currently, the classical Feldkamp-Davis-Kress (FDK) reconstruction algorithm~\cite{feldkamp1984practical} is considered the gold standard for reconstructing 4D DSA images in clinical practice. However, it requires acquiring a substantial number of 2D DSA images from different perspectives. This results in the generation of a significant amount of radiation, posing a threat to the health of the patient. Therefore, achieving better 4D DSA novel view synthesis results with fewer views is an important research issue. Moreover, real-time rendering speed plays a crucial role in enabling doctors to promptly visualize lesions and interact with them. Our research is primarily focused on developing 4D DSA novel view synthesis methods that can simultaneously achieve high rendering speed and superior rendering quality.

%The real-time rendering of 4D DSA helps observe the details of brain blood flow, which is of great significance for clinical visualization and interaction.

\iffalse
\begin{figure}[!h]
\centering
\hspace{0.42cm}\includegraphics[width=3.4in]{fig9.pdf}
\caption{Performance of different methods in 4D DSA novel view synthesis (30 training views).}
\label{fig5}
\vspace{-10pt}
\end{figure}
\fi

In recent years, neural radiance fields(NeRF)~\cite{mildenhall2021nerf, barron2021mip, zhang2020nerf++, barron2023zip} have achieved remarkable results in tasks such as novel view synthesis and scene reconstruction. It is also widely used in the field of medical imaging~\cite{Molaei_2023_ICCV}. As an implicit scene representation method, NeRF directly employs Multilayer Perceptrons (MLPs) to map the input position and direction information to color and density information. The resulting image is then obtained through volume rendering techniques~\cite{drebin1988volume}. 
However, due to the requirement of multiple MLP calculations for each rendered image, the original NeRF model incurs significant training and rendering time overhead. In recent studies, there have been various attempts to improve NeRF by using explicit representations instead of implicit representations~\cite{muller2022instant}\cite{xu2022point} or by employing plane decomposition techniques~\cite{chen2022tensorf}. These approaches aim to reduce memory usage and enhance training and rendering speed.

3D Gaussian Splatting~\cite{kerbl20233d} utilizes 3D Gaussians to represent the scene and employs an efficient differentiable splatting technique as a replacement for volume rendering. This approach enables real-time rendering while ensuring high-quality images~\cite{chen2024survey}. In addition to the advantage of fast rendering speed, 3D Gaussian splatting (3DGS) also avoids the computational overhead of rendering blank regions in space. Traditional NeRF requires sampling a large number of points in space, including blank regions. However, in 3DGS, the Gaussians only occur in non-blank regions, thus avoiding rendering blank areas. 4D DSA images often contain a significant amount of blank regions, making Gaussian splatting a suitable method for rendering such scenes. Since 4D DSA images are dynamic, it is necessary to find an appropriate way to extend 3D Gaussian splatting to handle the dynamics of the 4D DSA data.

Currently, several methods extend 3D Gaussian Splatting (3DGS) to dynamic scenes, One class of methods directly extends 3D Gaussians to 4D Gaussians~\cite{yang2023real, li2023spacetime, luiten2023dynamic}, while another class of methods models the temporal variations of Gaussians using deformable fields~\cite{yang2023deformable, wu20234d, liang2023gaufre, lin2023gaussian}. However, these works primarily focus on dynamic scenes in natural environments. The dynamic characteristics of 4D DSA data are different from those of natural scene data. As shown in Fig.~\ref{fig1}, in natural scenes, objects change position, shape, opacity, and color over time, with the most significant change being the displacement of objects. In the 4D DSA novel view synthesis task, the blood vessels in the brain remain stationary, and the signal intensity of the contrast agent in the blood vessels changes with the flow of blood, the primary variation occurs in the opacity of blood vessels. Therefore, the emphasis on reconstructing natural scenes and 4D DSA scenes differs. The current dynamic 3DGS methods are primarily designed for natural dynamic scenes. \textbf{When directly applying these methods to the 4D DSA novel view synthesis task, issues such as redundant module design, low computational efficiency, severe overfitting, and difficulties in achieving convergence may arise.}

\begin{figure*}[!h]
\centering
\hspace{-0.2cm}
\includegraphics[width=7.2in]{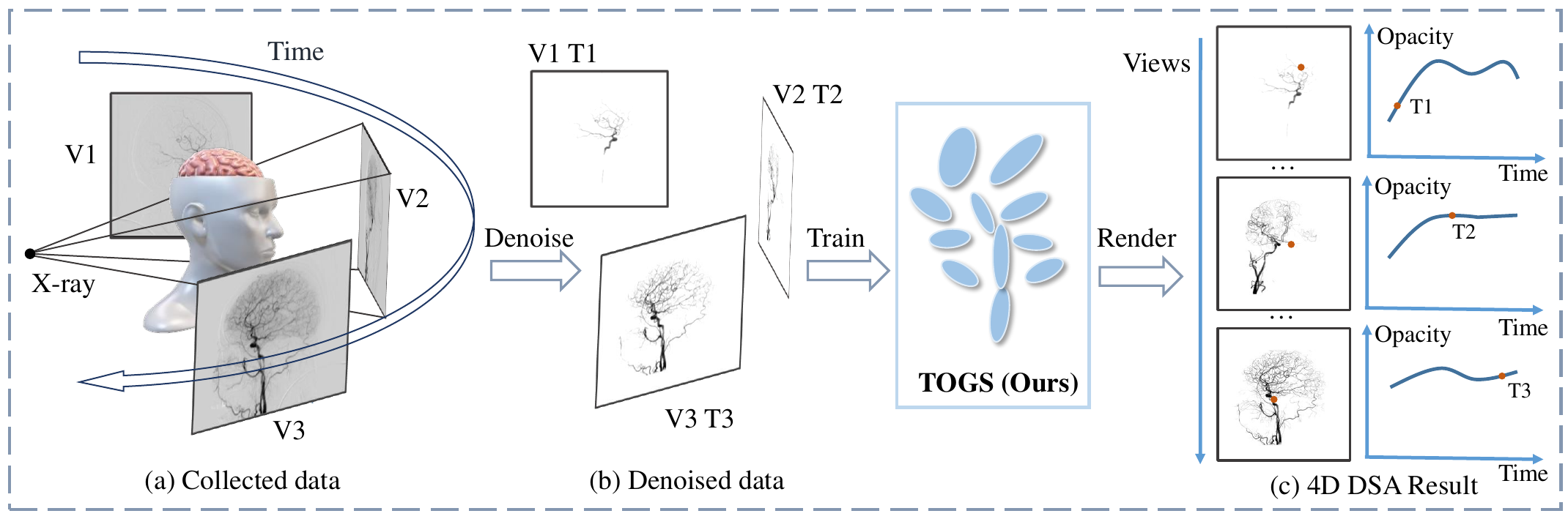}
\caption{An overview of the proposed TOGS for 4D DSA novel view synthesis. DSA data (a) is collected from different viewpoints and time points. The image is then denoised to retain only the blood vessel region. The denoised data (b) is subsequently used to train the TOGS model. The trained TOGS model is capable of rendering DSA images (c) from any desired time point and viewpoint. }
\label{fig3}
\vspace{-10pt}
\end{figure*}

To address the unique problem of 4D DSA novel view synthesis, we propose a simple and efficient method to extend 3DGS to the scene with changing opacity, specifically for modeling the diffusion process of the contrast agent within blood vessels. In this process, the blood vessels remain stationary, and only the signal intensity of the contrast agent varies. Therefore, in our approach, we only model the changes in opacity within the Gaussians. For each Gaussian, we introduce an opacity offset table to store the offset values of opacity at different times. As shown in Fig.~\ref{fig1} (d), the opacity of each Gaussian is obtained by querying or interpolating the opacity offset table and adding it to the original opacity value, Then get the rendered image by splatting. Our method is specifically designed for 4D DSA novel view synthesis. As shown in Fig.~\ref{fig2}. It offers advantages such as low memory consumption, fast rendering speed, and high rendering quality. Furthermore, the proposed method's use of opacity offset tables provides greater interpretability compared to the MLPs commonly used in previous methods. The entire process of 4D DSA novel view synthesis is depicted in Fig.~\ref{fig3}.

In summary, our contributions can be summarized as follows:

%(1) To the best of our knowledge, we are the first to introduce the Gaussian splatting technique to the 4D DSA reconstruction problem. By introducing an opacity offset table for each Gaussian, we extend 3DGS to the dynamic scenes of 4D DSA with relatively low time and space costs.

%(1) We introduce an effective  TOGS framework to handle the 4D DSA novel view synthesis problem, in which opacity offset tables are utilized to connect 3D Gaussians' status at different times.

(1) We introduced an effective TOGS framework to address the novel view synthesis problem in 4D DSA, which employs explicit opacity offset tables to model the time-varying opacity of 3D Gaussians.

(2) We introduce a Smooth loss to alleviate the overfitting problem that occurs in the sparse view case of the proposed method. We randomly prune Gaussians during the training process, which not only reduces memory overhead but also helps improve the quality of rendered images.

(3) Experiments prove that under the same number of views, the proposed method can achieve state-of-the-art rendering results. Additionally, it provides lower memory overhead and enables rendering speeds of over 300 FPS.

\section{Related Works}

\subsection{Dynamic scene reconstruction}
In recent years, NeRF~\cite{mildenhall2021nerf} has been widely used in the 3D reconstruction of static scenes. It implicitly represents the scene using MLPs. MLP maps the position information and perspective information of points in the 3D space into their color and density information and then obtains the image through volume rendering. The original NeRF method has high memory and time overhead. Therefore, subsequent research has focused on improving training and rendering speed, as well as reducing memory consumption~\cite{gao2022nerf}. These include explicit representations~\cite{fridovich2022plenoxels}\cite{sun2022direct} and tensor decomposition~\cite{chen2022tensorf} methods.

Currently, many works have extended NeRF for dynamic scene reconstruction, and a mainstream approach is to introduce a deformation field to model the spatial movement of points in the scene, such as Nerfies~\cite{park2021nerfies}, D-NeRF~\cite{pumarola2021d}, HyperNeRF~\cite{park2021hypernerf} and TiNeuVox~\cite{fang2022fast} etc. In addition, there are some methods 
\cite{cao2023hexplane, fridovich2023k, shao2023tensor4d} that use feature planes to capture 4D features to achieve high-quality dynamic new views synthesis and faster convergence speed. Although these methods achieve fast training speed, they are still challenging in real-time rendering of dynamic scenes.

The recent 3D Gaussian Splatting (3DGS)~\cite{kerbl20233d} adopts 3D Gaussian to represent the scene and utilizes efficient splatting techniques for image rendering. As a result, it significantly improves rendering speed and has gained widespread attention. There are multiple approaches to extend 3DGS to dynamic scenes~\cite{luiten2023dynamic, yang2023real, li2023spacetime, yang2023deformable, wu20234d, liang2023gaufre, lin2023gaussian}. Some methods directly allocate 3D Gaussians for each time point or introduce time attributes to Gaussians to represent dynamic scenes~\cite{luiten2023dynamic, yang2023real, li2023spacetime}. For example, Dynamic 3DGS~\cite{luiten2023dynamic} models dynamic scenes by tracking the position and variance of each Gaussian function at each timestep. This method has difficulties in accurately modeling monocular dynamic scenes, and it explicitly stores all the information of each Gaussian at each moment, leading to substantial storage overhead. Another popular extension method involves utilizing a deformation field to simulate the variations in 3D Gaussian attributes, such as position, shape, and opacity~\cite{yang2023deformable, wu20234d, liang2023gaufre, lin2023gaussian}. For example, In 4DGS~\cite{wu20234d}, multi-resolution neural voxels and a decoder MLP are used as the deformation field. It utilizes HexPlane to connect different adjacent Gaussian fields, aiming to predict more precise motion and shape deformations. The methods designed for natural dynamic scenes that model the displacement and shape changes of Gaussians may contain redundancy when applied to the 4D DSA novel view synthesis problem. Additionally, using MLPs or other deformation fields to model opacity transformations can introduce extra computational and memory overhead, while also reducing rendering speed. Hence, there is a requirement to develop an efficient Gaussian splatting method that is specifically designed for the 4D DSA novel view synthesis problem.

\subsection{4D DSA novel view synthesis}
Compared to 3D DSA reconstruction, 4D DSA provides temporal information that aids in visualizing abnormal features of certain diseases. For instance, it enables the visualization of internal characteristics of AVM lesions or the flow sequence of blood entering and leaving abnormal regions. Recent research~\cite{ruedinger2019optimizing}~\cite{chen2023estimating} suggests that the temporal information contained in 4D-DSA enables the quantification of blood flow velocity. When combined with accurate vascular measurements, it allows for the quantification of blood flow. Specifically, in 4D-DSA, each point has a time-density curve (TDC). By utilizing the time-density curves between two points, along with the distance between them and the cross-sectional area of the vessel, blood flow can be quantified.

Among the current reconstruction methods for 4D DSA images, the gold standard used in clinics is the classical Feldkamp-Davis-Kress (FDK) reconstruction algorithm~\cite{ feldkamp1984practical}. However, this algorithm requires a large number of 2D DSA images, resulting in significant exposure to radiation and potential harm to the patient. The emergence of deep learning methods has promoted research on CT/DSA image reconstruction~\cite{zhang2018sparse, han2018framing, zhao2022self, zhang2020metainv}. These methods are mainly aimed at static CT reconstruction and do not consider the flow of contrast agents.

In recent years, NeRF has been increasingly applied to medical image reconstruction tasks, including CT reconstruction~\cite{zha2022naf}~\cite{corona2022mednerf} and deformable tissue reconstruction~\cite{wang2022neural}~\cite{yang2023neural}. In EndoNeRF~\cite{wang2022neural}, it follows the modeling in D-NeRF~\cite{pumarola2021d} and represents deformable surgical scenes as a canonical neural radiance field along with a time-dependent neural displacement field. In LerPlane~\cite{yang2023neural}, it treats surgical procedures as 4D volumes and factorizes them into explicit 2D planes of static and dynamic fields, leading to a compact memory footprint and significantly accelerated optimization. Specifically addressing the 4D DSA novel view synthesis problem, TiAVox~\cite{zhou2023tiavox} introduces a 4D attenuation voxel grid to reflect the attenuation characteristics of 4D DSA from both spatial and temporal dimensions. TiAVox has achieved shorter training times. However, there is still significant room for improvement in terms of the rendering speed of the model.

3DGS achieves real-time rendering while maintaining high rendering quality, which greatly enhances visualization. Paper~\cite{li2023sparse} applied 3DGS to sparse CT reconstruction and achieved good performance. Therefore, it is necessary to introduce Gaussian splatting techniques into the 4D DSA novel view synthesis problem. Existing 4D Gaussian splatting methods mainly focus on dynamic scenes in the natural world, which differ from the dynamic processes in 4D DSA. Hence, this paper designs a Gaussian splatting algorithm specifically for the 4D DSA novel view synthesis problem. We directly employ an opacity offset table to model the opacity offset of Gaussians over time. Compared to current 4DGS methods, this approach offers advantages such as lower computational requirements, faster rendering speed, and smaller memory overhead.

\begin{figure*}[!h]
\centering
%\hspace{0.42cm}
\includegraphics[width=7.2in]{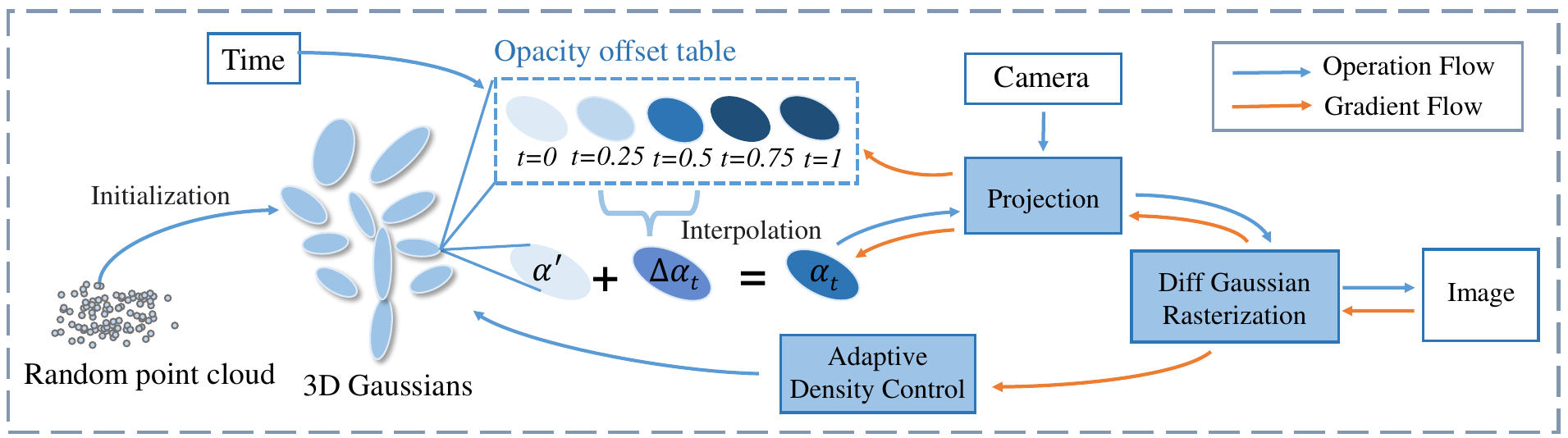}
\caption{Overview of our pipeline. First, we randomly generate point clouds and initialize them as 3D Gaussians. We introduce an opacity offset table for each Gaussian. When different time values $t$ are input, interpolate the opacity offset table to obtain the opacity offset value $\Delta \alpha_{t}$. Add it to the original opacity $\alpha^{\prime}$ to obtain the opacity $\alpha_{t}$ at that moment. Then, Gaussians are input into the efficient differential Gaussian rasterization pipeline to render the image. Finally, by computing the loss and backpropagating gradients, we adaptively control the density of the Gaussians.}
\label{fig4}
\vspace{-10pt}
\end{figure*}

\vspace{-5pt}
\section{Preliminary}
\subsection{3D Gaussian Splatting}

3DGS\cite{kerbl20233d} represents the scene through a large number of 3D Gaussians, and each Gaussian is characterized by a covariance matrix $\Sigma \in \mathbb{R}^{3\times 3} $ and mean $\mu \in \mathbb{R}^3$:
\begin{equation}
G(X)=e^{-\frac{1}{2} (x-\mu^T) \Sigma^{-1} (x-\mu)},
\end{equation}
where the covariance matrix $\Sigma$ is divided into a quaternion matrix $R$ representing rotation and a matrix $S$ representing scaling:
\begin{equation}
\Sigma = RSS^{T}R^{T}.
\end{equation}

Paper\cite{zwicker2001surface} demonstrates how to project a 3D Gaussian into 2D. Given the viewing transformation $W$ and the Jacobian of the affine approximation of the projective transformation $J$, the covariance matrix $\Sigma^{\prime}$ in camera coordinates can be obtained by:
\begin{equation}
\Sigma^{\prime}=J W \Sigma W^T J^T.
\end{equation}

In summary, each Gaussian $G(x,r,s,\alpha,c)$ is represented by the following attributes: position $x \in \mathbb{R}^3$, opacity $\alpha \in \mathbb{R}$, rotation factor $r \in \mathbb{R}^4$, and scaling factor $s \in \mathbb{R}^3$, and the color represented by the spherical harmonics $c$.
To render a 2D image, 3DGS sorts all the Gaussians that make up a pixel and determines the color of each pixel by:

\begin{equation}
C=\sum_{i=1}^{n} c_i \sigma_i \prod_{j=1}^{i-1}\left(1-\sigma_j\right),
\end{equation}
where the $c_i$ is the color calculated from the SH coefficient of the $i$-th Gaussian function. The $\sigma_i$ is obtained by multiplying the opacity $\alpha$ with the covariance $\Sigma^{\prime}$ of the two-dimensional Gaussian function:
\begin{equation}
\sigma_i=\alpha_i\times e^{-\frac{1}{2} (x^{\prime}-\mu_i^{\prime})^T\Sigma^{-1} (x^{\prime}-\mu_i^{\prime})},
\end{equation}
where $x$ and $\mu_i$ are coordinates in the projected space.

3DGS uses an adaptive density control strategy to control the number of Gaussians. For the under-reconstructed area, the Gaussian is copied and moved along the direction of the positional gradient. On the other hand, the large Gaussian distribution in the high variance area is divided into small Gaussian distributions. In addition, it deletes Gaussian whose opacity is less than the threshold every certain iteration.

\section{Methods}
\subsection{The overview of the proposed model}

The pipeline of the proposed model is illustrated in Fig.~\ref{fig4}. Generate a set of point clouds through random initialization, and use the point cloud to generate a set of 3D Gaussians $G(x,r,s,\alpha,c)$ defined by a center
position $x$, opacity $\alpha$, and 3D covariance matrix $\Sigma$ obtained
from quaternion $r$ and scaling $s$. In order to model the signal intensity changes of contrast agents in 4D DSA, we introduce an opacity offset table $T_{\alpha}$ to each Gaussian, resulting in $G(x,r,s,\alpha, T_{\alpha},c)$. After the current time $t$ is input, the opacity offset value $\Delta \alpha_t$ is obtained by interpolating the opacity offset table. Then $G(x,r,s,\alpha+\Delta \alpha_t,c)$ is sent to the efficient differential Gaussian rasterization pipeline to render the image. By calculating the loss and performing backpropagation, we apply adaptive density control to the Gaussians.

\subsection{Opacity offset table}
In the 4D DSA novel view synthesis problem, the blood vessels remain stationary, while the signal intensity of the contrast agent inside the vessels varies due to blood flow. Therefore, the proposed method specifically focuses on modeling the temporal changes in Gaussian opacity. We directly introduce an opacity offset table $T_{\alpha}$ for each Gaussian to store the opacity offset values at different time points, so as to reduce additional calculation and space overhead as much as possible. During the rendering process, the opacity of the Gaussian at a given moment is calculated by adding the initial opacity value to the opacity offset value obtained through linear interpolation from the opacity offset table $T_{\alpha}$.

\begin{equation}
\alpha_{t}=\alpha^{\prime}+\Delta \alpha_{t},where\ \Delta \alpha_{t} = interp(T_{\alpha}),
\end{equation}
where $\alpha^{\prime}$ is the initial opacity of the Gaussian, and  $\Delta \alpha_{t}$ is the opacity offset value of the Gaussian at time $t$. In this paper, we use an opacity offset table of length 5 to store the opacity offset values for Gaussian at time $t$ equal to 0, 0.25, 0.5, 0.75, and 1, respectively.

\subsection{The initialization and density control of Gaussians}

\subsubsection{Random initialization} 
Generating a large number of effective initial point clouds using Structure from Motion (SfM)~\cite{schonberger2016structure} can be challenging in the case of 4D DSA, as it is a dynamic imaging technique with only one image available for each moment and viewpoint. Therefore, this paper randomly generates 100,000 point clouds as initialization.

\subsubsection{Pruning Gaussians based on maximum opacity}One of the key reasons why 3D GS can effectively represent the scene is the adoption of Gaussian adaptive density control. This strategy sets an opacity threshold and removes Gaussians with opacity below the threshold every certain number of iterations. During the 4D DSA imaging process, many points initially have low opacity, and their opacity gradually increases over time. In this paper, we prune the Gaussian when the maximum opacity of the Gaussian is below the threshold value $U$, as shown in the following formula:
\begin{equation}
\alpha^{\prime}+max(\Delta \alpha_{t})<U.
\label{equation6}
\end{equation}

\subsubsection{Randomly pruning Gaussians}Due to the utilization of random initialization and the inherent characteristics of 3DGS, the model is prone to generating an excessive number of Gaussians when fitting scenes under sparse viewing angles, which can result in overfitting issues. During the training phase of the model, Gaussian splitting, duplication, and pruning are executed between the 1000th and 20000th iterations. In this particular stage, for every 200 iterations, an extra 8\% of the Gaussians are randomly pruned. This strategy allows us to eliminate redundant Gaussians, decrease the memory usage of the model, and enhance the quality of the rendered output simultaneously.

\subsection{Optimization}

The loss function of the proposed model consists of three components: Smooth loss, reconstruction loss, and Structural Similarity Index Measure (SSIM)~\cite{wang2004image} loss.

\subsubsection{Smooth loss}We introduced an opacity offset table to simulate the signal intensity variation of contrast agent flow. However, during the experiments, we observed the presence of overfitting issues. To address this, we proposed a Smooth loss to constrain the optimization of the opacity offset table:

\begin{equation}
L_{\text {Smooth }}=\frac{1}{N} \sum_{k=0}^N \sum_{t=1}^L\left|\Delta \alpha_{t}^k-\Delta \alpha_{t-1}^k\right|,
\end{equation}
where $\Delta \alpha_{t}^k$ represents the value indexed as $t$ in the opacity offset table of the $k$-th Gaussians, $N$ represents the total number of points, and $L$ represents the length of opacity offset table.

By introducing the Smooth loss, we reduce the differences between consecutive opacity offsets in the opacity offset table. This promotes a smoother temporal variation of opacity for the Gaussians, which helps alleviate overfitting issues.

\subsubsection{Reconstruction loss}Similar to 3DGS~\cite{kerbl20233d}, we introduce the reconstruction loss in the model:
\begin{equation}
L_{\text {recon }}=\|\hat{I}-I\|_1,
\end{equation}
where $\hat{I}$ represents the target image, and $I$ represents the rendered image.

\subsubsection{SSIM loss}We introduce SSIM loss to improve the quality of generated images:
\begin{equation}
L_{S S I M}=1-\operatorname{SSIM}(\hat{I}, I).
\end{equation}

Therefore, the final loss function is:
\begin{equation}
L_{\text {total }}=L_{\text {recon }}+\lambda_{\text {SSIM }} L_{\text {SSIM }}+\lambda_{\text {Smooth }} L_{\text {Smooth }},
\end{equation}
 where $\lambda_{\text {SSIM }}$ represents the weight of SSIM loss, and  $\lambda_{\text {Smooth}}$ represents the weight of Smooth loss.

\section{Experiment}

\subsection{Implementation Details}

We train our model using the Adam\cite{kingma2014adam} optimizer, with a total of 30,000 iterations for each experiment. The learning rates for position, opacity, size, and rotation parameters of the Gaussian are set as follows: $1.6\times10^{-4}$, 0.025, 0.005, and 0.001, respectively. The hyperparameters of the loss function are tuned based on the number of training viewpoints. In the case of 30 views, the values of the hyperparameters are as follows: $\lambda_{\text {SSIM }}$ is 0.1, $\lambda_{\text {Smooth }}$ is 0.0043, and the opacity threshold $T$ is set to 0.018. The proposed method is implemented in Python, using the PyTorch deep learning framework. All experiments are conducted on a single NVIDIA RTX 3090 GPU with 24GB memory.

\iffalse
\begin{figure}[!t]
\centering
\hspace{0.42cm}\includegraphics[width=3.6in]{fig3.pdf}
\vspace{-20pt}
\caption{An example of (a) Clinically collected, (b) Mask, and (c) Denoised images.}
\label{fig3}
\vspace{-10pt}
\end{figure}
\fi

\subsection{Datasets}
We use data provided by TiAVox~\cite{zhou2023tiavox}, collected from 8 patients in Wuhan Union Hospital. For each patient's data, a total of 133 2D projection images were captured. Each image was acquired at different time points and perspectives. We normalize the time corresponding to the images from 0 to 1. The resolution of the image is 1024$\times$1024 pixels.

Clinical DSA images often contain various sources of noise, such as bone signals, imaging artifacts, and other interferences. These factors significantly affect the quality of synthesized new views. As 4D DSA novel view synthesis focuses specifically on the vascular region, a Unet is used in TiAVox\cite{zhou2023tiavox} to segment the blood vessel area and generate mask images of the blood vessel. By applying mask images, we obtain clear images containing only the blood vessel area, as shown in Fig.~\ref{fig3} (b).

\label{sec:guidelines}

\subsection{Evaluation metrics}

In this paper, we use Peak Signal Noise Ratio (PSNR), SSIM, and Learned Perceptual Image Patch Similarity (LPIPS)~\cite{zhang2018unreasonable} as the three metrics to evaluate the quality of the generated images by the model in the task of synthesizing new views. 
\iffalse
Among them, the PSNR is calculated by:

\begin{equation}
PSNR(x,y)=10\cdot log_{10}(\frac{MAX_{I}^{2}}{MSE(x,y)} ),
\end{equation}
where $MAX(I)$ represents the maximum possible pixel value in the image, and $MSE(x,y)$ represents the mean square error of the two images.

The SSIM indicator is calculated by the following formula:
\begin{equation}
SSIM(x,y)=\frac{(2\mu _{x}\mu _{y}+C_{1})(2\sigma _{xy}+C_{2})}{(\mu _{x}^{2}+\mu_{y}^{2}+C_{1})(\sigma_{x}^{2}+\sigma_{y}^{2}+C_{2})},
\end{equation}
where $\mu _{x}$ and $\mu _{y}$ represent the mean of the image $x$ and $y$ respectively, $\sigma_{x}$ and $\sigma_{y}$ represent the standard deviation of the two images respectively, and $\sigma _{xy}$ represents the covariance of the two images.

The score of LPIPS is determined by the weighted per-pixel mean squared error (MSE) of multiple layers of feature maps:

\begin{equation}
LPIPS(x,y)=\sum_{l}^{L} \frac{1}{H_{l}W_{l}} \sum_{h,w}^{H_{l},W_{l}}\left \| w_{l}\odot (x_{0hw}^{l}-y_{0hw}^{l}) \right \|_{2}^{2},
\end{equation}
where $x_{0hw}^{l}$, $y_{0hw}^{l}$ are the reference and assessed images’ features at pixel width $w$, pixel height $h$, and layer $l$. $H_{l}$ and $W_{l}$ are the feature map's height and width at the corresponding layer. Scale the $x_{0hw}^{l}$ and $y_{0hw}^{l}$ by vector $w_{l}$ and calculate the $L2$ distance. 
\fi
We use the VGG network~\cite{simonyan2014very} as the feature extraction backbone to calculate LPIPS.
\vspace{-5pt}

\begin{figure*}[!h]
\centering
\hspace{0.4cm}\includegraphics[width=5.9in]{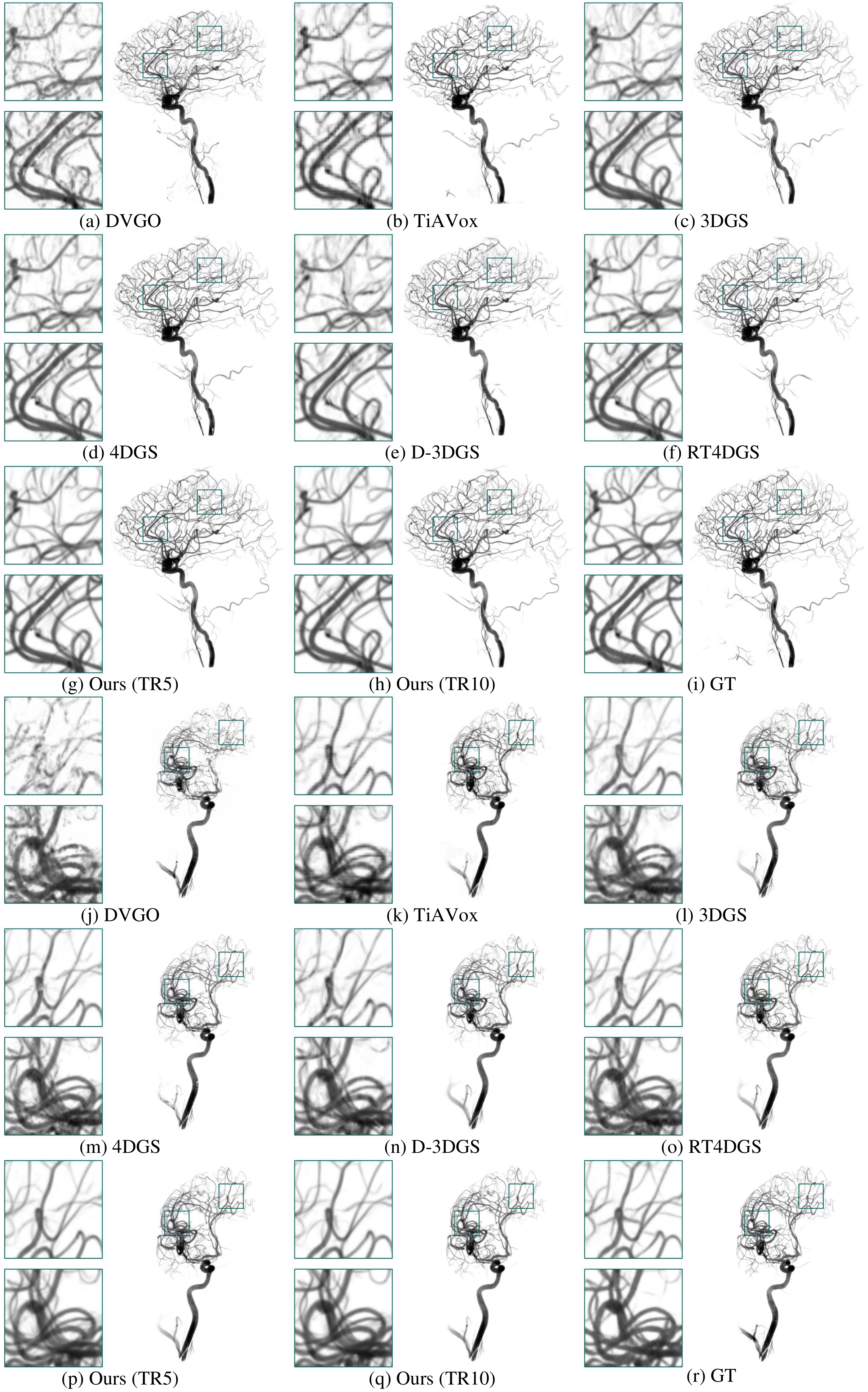}
\vspace{-5pt}
\caption{Visualization comparison results of different methods in the 4D DSA new view synthesis task. (Number of training and testing views: 30/103, TR: Temporal resolution).}
\label{fig5}
\vspace{-10pt}
\end{figure*}

\subsection{Comparison with Existing Methods}
We conducted experiments on novel view synthesis when the number of training views was 10, 20, 30, 40, 50, and 60, respectively. We selected DVGO~\cite{sun2022direct}, TiAVox~\cite{zhou2023tiavox}, 3DGS~\cite{kerbl20233d}, D-3DGS~\cite{yang2023deformable}, RT4DGS~\cite{yang2023real} and 4DGS~\cite{wu20234d} as the six methods for comparison in experiments. Among them, DirectVoxGo and TiAVox use voxel grids to represent scenes, and 3DGS, D-3DGS, RT4DGS and 4DGS use 3D Gaussians to represent scenes. We found that modeling position offset in 4D DSA can potentially hinder model convergence. Therefore, in our comparative experiment, we removed the position offset of Gaussians modeled in 4DGS and D-3DGS. For detailed experimental information, please refer to the \ref{experiment} section of the paper.

The results of the comparative experiment are shown in Fig.~\ref{fig5} and Table~\ref{tab1}. We reported the experimental results of the model at temporal resolutions (length of the opacity offset table) of 5 and 10. From Fig.~\ref{fig5}, it is evident that the reconstruction results of DVGO exhibit noticeable noise and fail to effectively capture the intricate details of the blood vessels. TiAvox is capable of recovering the shape of the vessels, however, the opacity of the vessels appears less realistic, and the vessel details are relatively blurry. As shown in Fig.~\ref{fig5} (l) and (m), in regions with a significant overlap of blood vessels, the results obtained from 3DGS and 4DGS appear cluttered, and the background is not adequately clean. In 4D DSA novel view synthesis, the opacity of many Gaussians is initially low and gradually increases over time. 4DGS directly clipping Gaussians based on their initial opacity consequently limits its reconstruction results. In D-3DGS and RT4DGS, the grayscale of blood vessels is not smooth enough and the rendered images are not clear enough. In contrast, our proposed method accurately recovers the details of the blood vessels while generating more realistic results. From Table~\ref{tab1}, it is evident that our proposed method outperforms existing approaches in terms of evaluation metrics, particularly in the case of sparse viewpoint quantities (10 views). Our method produces higher-quality novel view images.

\begin{table*}[t!]
    \centering
    \caption{Comparison of new view synthesis results of different methods (Total number of views:133, TR: Temporal resolution)}
    \label{tab1}
    \begin{tabular}{ccccccccccc}
        \toprule
        \multirow{2}{*}{TRAIN VIEWS}  & \multicolumn{3}{c}{10} & \multicolumn{3}{c}{20} & \multicolumn{3}{c}{30} & \\
          & PSNR$\uparrow$ & SSIM$\uparrow$ & LPIPS$\downarrow$  & PSNR$\uparrow$  & SSIM$\uparrow$  & LPIPS$\downarrow$  & PSNR$\uparrow$ & SSIM$\uparrow$ & LPIPS$\downarrow$ \\
        \midrule
        %NAF~\cite{zha2022naf}      & 26.06	 & 0.85 && 26.78 & 0.88	 &&26.79& 0.90 &    \\
        DirectVoxGo~\cite{sun2022direct}  & 24.85 & 0.905 & 0.115&26.86 & 0.917 &0.108& 28.23 & 0.933& 0.098   \\
        TiAVox~\cite{zhou2023tiavox}  & 28.57 & 0.939 &0.091& 30.46 & 0.951 & 0.077& 31.23 & 0.959& 0.065  \\
        3DGS~\cite{kerbl20233d} & 27.23 & 0.924 &0.095 &29.82 & 0.947  &0.063& 31.25 & 0.961& 0.050 \\
        D-3DGS~\cite{yang2023deformable}  & 26.08 & 0.913 &0.108 & 28.43 & 0.932 &0.083 & 30.58 & 0.953 &0.067\\
        4DGS~\cite{wu20234d}  & 27.36 & 0.927 &0.089& 30.64 & 0.953	 &0.058& 31.53 & 0.962&0.050  \\
        RT4DGS~\cite{yang2023real}& 29.43 & 0.948 & 0.063& 31.12 & 0.957 	 & 0.053 & 31.82 & 0.965& 0.047  \\
        TOGS (TR5)  & 29.83 & \textbf{0.949}&0.063&31.48 & 0.959 &0.050& 32.28& 0.969 & 0.041\\
        TOGS (TR10)  & \textbf{29.94} & 0.948 &\textbf{0.063} &\textbf{31.65} & \textbf{0.960} &\textbf{0.049}
        & \textbf{32.47} & \textbf{0.969} &\textbf{0.040} \\
        \midrule
        \multirow{2}{*}{TRAIN VIEWS}  & \multicolumn{3}{c}{40} & \multicolumn{3}{c}{50} & \multicolumn{3}{c}{60} & \\
          & PSNR$\uparrow$ & SSIM$\uparrow$ & LPIPS$\downarrow$  & PSNR$\uparrow$  & SSIM$\uparrow$  & LPIPS$\downarrow$  &
          PSNR$\uparrow$  & SSIM$\uparrow$  & LPIPS$\downarrow$  &\\
        \midrule
        %NAF~\cite{zha2022naf}    & 26.89 & 0.90 && 27.07 & 0.91	 &  &   \\
        DirectVoxGo~\cite{sun2022direct} & 29.18 & 0.940 &0.090& 29.82 & 0.942	 &0.086  &30.75&0.951&0.077   \\
        TiAVox~\cite{zhou2023tiavox} & 31.51 & 0.962 &0.063& 31.94 & 0.962	 &0.065  & 32.29&0.965& 0.062 \\
        3DGS~\cite{kerbl20233d}& 31.87 & 0.965 & 0.048& 32.59 & 0.968&0.040  &33.18&0.972 &0.036 \\
        D-3DGS~\cite{yang2023deformable} &30.67 & 0.955 &0.064 &30.83 & 0.956 &0.063 &31.40 & 0.961 &0.057\\
        4DGS~\cite{wu20234d}& 31.93 & 0.965 & 0.049& 32.71 & 0.970 	 & 0.039 & 33.22 & 0.974 & 0.034   \\
        RT4DGS~\cite{yang2023real}&32.12 & 0.967& 0.047& 32.35 & 0.967& 0.046 & 32.86 & 0.971& 0.042   \\
        TOGS (TR5)  & 32.60 & 0.970 &0.040 &33.36 & 0.973&0.035& 33.81 & 0.975 &0.033  \\
        TOGS (TR10)  & \textbf{32.82} & \textbf{0.971} &\textbf{0.039} &\textbf{33.49} & \textbf{0.973} &\textbf{0.034}
        & \textbf{33.96} & \textbf{0.976} &\textbf{0.032}  \\
        
        \bottomrule
    \end{tabular}

    \vspace{-10pt}
\end{table*}

\begin{table}[htbp]
    \centering
    \caption{Comparison of Model Speed and Storage Occupancy (30 training views)}
    \label{tab2}
    \begin{tabular}{cccc}
        \toprule
        \multirow{2}{*}{Method}  & Render speed$\uparrow$ & Train time$\downarrow$  &Storage$\downarrow$\\
         & (FPS)  & (minute) &(MB)\\
        \midrule
        %NAF     & -	& \textgreater30  \\
        DirectVoxGo~\cite{sun2022direct}      & 1.72  &21.1 & 1650\\
        TiAVox~\cite{zhou2023tiavox}  & 1.46 &\textbf{4.9} &  901\\
        3DGS~\cite{kerbl20233d}      & \textbf{384} &5.0 & 80 \\
        D-3DGS~\cite{yang2023deformable}  & 266 &10.0 & \textbf{12} \\
        4DGS~\cite{wu20234d}   & 75 &20.0 & 172\\ 
        RT4DGS~\cite{yang2023real} &101 &42.0 & 82\\
        TOGS (Ours)  & 317&6.0 & 13\\

        \bottomrule
    \end{tabular}
    \vspace{-5pt}
\end{table}

%it can be observed that our proposed method significantly outperforms existing methods in terms of the evaluated metrics, achieving high-quality synthesis of new viewpoints.

We also conducted a comparison of rendering speed, training time, and storage size among different methods, as shown in Table~\ref{tab2}. The Gaussian Splatting-based method exhibits significantly faster rendering speed compared to the voxel grid-based method. Our method extends 3DGS to dynamic scenes in 4D DSA while maintaining a high rendering speed. Regarding training time, our model can complete the training process within 6 minutes, maintaining a high training speed. In terms of storage overhead, the proposed model achieves the minimum storage cost. This is because the proposed model can effectively represent the scene using only a small number of Gaussians, without relying on any additional neural networks.

\iffalse
\begin{figure}[!h]
\centering
\hspace{0.42cm}\includegraphics[width=3.4in]{fig9.pdf}
\caption{Performance of different methods in 4D DSA novel view synthesis (30 training views).}
\label{fig5}
\vspace{-10pt}
\end{figure}
\fi

\begin{table*}[h!]
    \centering
    \caption{Ablation study for opacity offset table and randomly pruning gaussians (30 training views)}
    \label{table3}
    \begin{tabular}{ccccccccccc}
        \toprule
        \multirow{2}{*}{VIEWS}  & \multicolumn{3}{c}{10}  & \multicolumn{3}{c}{30}  & \multicolumn{3}{c}{50} \\
          & PSNR$\uparrow$ & SSIM$\uparrow$   & LPIPS$\downarrow$  & PSNR$\uparrow$ & SSIM$\uparrow$& LPIPS$\downarrow$  & PSNR$\uparrow$  & SSIM$\uparrow$& LPIPS$\downarrow$   \\
        \midrule
        w/o opacity offset table  &  28.98	& 0.945 &0.066   &  31.67	& 0.965 &0.047  & 32.67 & 0.970 & 0.039\\
        w/o randomly pruning Gaussians  & 29.72 & 0.946&0.064 & 32.16 & 0.967 &0.042 & 33.13 & 0.971 &0.037\\
        all  & \textbf{29.83} & \textbf{0.949} &\textbf{0.063} & \textbf{32.28}& \textbf{0.969} & \textbf{0.041} & 
               \textbf{33.36} & \textbf{0.973} &\textbf{0.035}\\

        \bottomrule
    \end{tabular}
    \vspace{-10pt}
\end{table*}

\begin{table*}[h!]
    \centering
    \caption{Ablation study for Loss function (30 training views)}
    \label{table4}
    \begin{tabular}{ccccccccccc}
        \toprule
        \multirow{2}{*}{VIEWS}  & \multicolumn{3}{c}{10}  & \multicolumn{3}{c}{30}  & \multicolumn{3}{c}{50} \\
          & PSNR$\uparrow$ & SSIM$\uparrow$   & LPIPS$\downarrow$  & PSNR$\uparrow$ & SSIM$\uparrow$& LPIPS$\downarrow$  & PSNR$\uparrow$  & SSIM$\uparrow$& LPIPS$\downarrow$   \\
        \midrule
        w/o Smooth loss  & 
        {29.27} & {0.943}&{0.066} & 
        {32.12} & {0.968} &{0.041} 
        & {33.33} & {0.973}& {0.035} \\
        w/o SSIM loss  & 
        {29.73} & {0.948}&{0.067} & 
        {31.76} & {0.964} &{0.049} 
        & {32.84} & {0.969}& {0.042} \\
        all  & \textbf{{29.83}} & \textbf{{0.949}} &\textbf{{0.063}} & \textbf{{32.28}}& \textbf{{0.969}} & \textbf{{0.041}} & 
               \textbf{{33.36}} & \textbf{{0.973}} &\textbf{{0.035}} \\
        \bottomrule
        \vspace{-10pt}
    \end{tabular}
\end{table*}

\vspace{-5pt}
\begin{table*}[h!]
    \centering
    \caption{Ablation study for pruning strategy (30 training views)}
    \label{table5}
    \begin{tabular}{cccccccccccccc}
        \toprule
        \multirow{2}{*}{Threshold $T$}  & \multicolumn{3}{c}{0.0005}& \multicolumn{3}{c}{0.001}  & \multicolumn{3}{c}{0.01}  & \multicolumn{3}{c}{0.015} \\
          & PSNR$\uparrow$ & SSIM$\uparrow$   & LPIPS$\downarrow$& PSNR$\uparrow$ & SSIM$\uparrow$   & LPIPS$\downarrow$  & PSNR$\uparrow$ & SSIM$\uparrow$& LPIPS$\downarrow$  & PSNR$\uparrow$  & SSIM$\uparrow$& LPIPS$\downarrow$   \\
        \midrule
        Initial opacity & {30.36}& {0.955} & {0.062} & {30.08}& {0.954} & {0.065} & {30.26}& {0.954} & {0.064}&{30.22}& {0.954} & {0.064}\\
        Mean opacity & {32.27}& {0.969} & {0.041} & {32.25}& {0.968} & {0.041} & {32.16}& {0.968} & {0.043}&{32.03}& {0.967} & {0.045} \\
        Max opacity &\textbf{{32.28}}& \textbf{{0.969}} & \textbf{{0.041}} &\textbf{{32.26}}& \textbf{{0.969}} & \textbf{{0.041}}& \textbf{{32.27}}& \textbf{{0.969}} & \textbf{{0.041}}& \textbf{{32.25}}& \textbf{{0.968}} & \textbf{{0.042}} \\
        \bottomrule
    \end{tabular}
    \vspace{-10pt}

\end{table*}

\iffalse
\begin{table}[h!]
    \centering
    \caption{The experimental results at different Temporal resolutions.}
    \label{tab6}
    \begin{tabular}{ccccc}
        \toprule
        VIEWS  &Temporal resolution   & PSNR$\uparrow$ & SSIM$\uparrow$  &LPIPS$\downarrow$\\
        \midrule
        \multirow{6}{*}{10} 
        &{0}    & {28.98}	& {0.945} &{0.066}  
        \\
        &{2}    & {29.46}	& {0.947} &{0.065}
        \\
        &{3}    & {29.74}	& {0.949} &{0.064}
        \\
        &{5}    & {29.83}	& {0.949} &{0.063}
        \\
        &{7}    & {29.93}	& \textbf{{0.949}}&\textbf{{0.062}}
        \\
        &{10}  & \textbf{{29.94}} & {0.948} &{0.063}
        \\
        \midrule

        \multirow{6}{*}{30} 
        &{0}    & {31.67}	& {0.965} &{0.047} 
        \\
        &{2}    & {31.92}	& {0.966} &{0.045} \\
        &{3}    & {32.09}	& {0.967} &{0.043}
        \\
        &{5}    & {32.28}& {0.969} & {0.041}
        \\
        &{7}    & {32.30}	& {0.969} &{0.041}
        \\
        &{10}  & \textbf{{32.47}} &\textbf{{0.969}} &  \textbf{{0.040}} 
        \\
        \bottomrule
    \end{tabular}

    \vspace{-5pt}
\end{table}
\fi

\begin{table}[h!]
    \centering
    \caption{The experimental results at different Temporal resolutions. (TR: Temporal resolution)}
    \label{tab6}
    \begin{tabular}{cccccc}
        \toprule
        VIEWS  &TR   & PSNR$\uparrow$ & SSIM$\uparrow$  &LPIPS$\downarrow$ & Gaussians Num\\
        \midrule
        \multirow{6}{*}{10} 
        &{0}    & {28.98}	& {0.945} &{0.066}   &{13315}
        \\
        &{2}    & {29.46}	& {0.947} &{0.065}  &{15016}
        \\
        &{3}    & {29.74}	& {0.949} &{0.064}  &{12120}
        \\
        &{5}     & {29.83}	& {0.949} &{0.063}  &{10909}
        \\
        &{7}    &  {29.93}	& \textbf{{0.949}}&\textbf{{0.062}}  &{9558}
        \\
        &{10}  & \textbf{{29.94}} & {0.948} &{0.063} &{9285}
        \\
        \midrule

        \multirow{6}{*}{30} 
        &{0}    & {31.67}	& {0.965} &{0.047} & {85669}
        \\
        &{2}    & {31.92}	& {0.966} &{0.045} & {37319}\\
        &{3}    & {32.09}	& {0.967} &{0.043} & {35073}
        \\
        &{5}    & {32.22}	& {0.968} &{0.042} & {31949}
        \\
        &{7}    & {32.30}	& {0.969} &{0.041} & {30531}
        \\
        &{10}  & \textbf{{32.47}} &\textbf{{0.969}} &  \textbf{{0.040}}  & {29460}
        \\
        \bottomrule
    \end{tabular}

    \vspace{-5pt}
\end{table}

\subsection{Ablation study}
\subsubsection{Modeling position offset experiment}
\label{experiment}

To verify whether it is suitable to model the displacement of objects in the 4D DSA novel view synthesis problem, we conducted experiments on two dynamic 3DGS methods, Deformable 3DGS~\cite{yang2023deformable}, and 4DGS, that both use deformable fields to model object displacements. In Deformable 3DGS,  we introduced a coefficient $w$ for position displacement $\delta x$. When $w$ is set to 0, the model generates a large number of point clouds, making it challenging to discern the shape of the blood vessels clearly from these point clouds. The model can render 2D DSA images normally. When $0<w<=0.3$, the point cloud shape is abnormal, but 2D DSA images can be rendered. When $w>0.3$, most of the point cloud is clipped, leaving only very few points, and 2D DSA images cannot be rendered normally. In 4DGS, when we remove the position displacement $\delta x$ of the point cloud, the point cloud shape can be well generated and the DSA image can be rendered at the same time. When the position offset is introduced, the shape of the blood vessels is partially missing, and the image cannot be rendered normally. This experiment confirms that modeling positional offset in 4D DSA novel view synthesis may prevent the model from converging properly, thereby affecting the rendering of images. When compared to 4DGS, Deformable 3DGS, and RT4DGS, we removed the positional offset modeling from these methods.

\subsubsection{Ablation study for Opacity offset table and Randomly pruning Gaussians}
We conducted ablation experiments to verify the effectiveness of the proposed model. First, we verified the effectiveness of the introduced opacity offset table and random pruning strategy, as shown in Table~\ref{table3}. After removing the opacity offset table, the model is still capable of fitting new viewpoint images, but the quality of the rendered images significantly deteriorates. Simultaneously, it can be observed from the table that randomly pruning a certain proportion of Gaussians can also enhance the quality of rendered images. Therefore, we conducted further research on the impact of randomly pruning Gaussians at different proportions on memory usage and the quality of the generated images. The experimental results are shown in Fig.~\ref{fig6}. We compared the experimental results with pruning proportions of 4\%, 8\%, 10\%, and 12\%. Taking 8\% as an example, we observed that compared to the results without random pruning strategy, the number of Gaussians decreased from 120,000 to 40,000, the model memory reduced from 33M to 11M, and the PSNR of the generated images improved from 31.69 to 31.81. This phenomenon suggests that in Gaussian splatting, the random initialization and Gaussian splitting duplication strategies may lead to the generation of excessive Gaussians to fit the scene, resulting in overfitting issues. Randomly pruning a certain proportion of Gaussians can help alleviate overfitting problems and reduce memory overhead.

\iffalse
\begin{figure}[!h]
\centering
\hspace{0.42cm}\includegraphics[width=3.4in]{fig6.pdf}
\caption{The impact of randomly pruning Gaussians at different proportions on the number of Gaussians and the quality of generated images (30 training views).}
\label{fig6}
\vspace{-10pt}
\end{figure}
\fi

\begin{figure}[!h]
\centering
\subfigure{\includegraphics[width=3.1in]{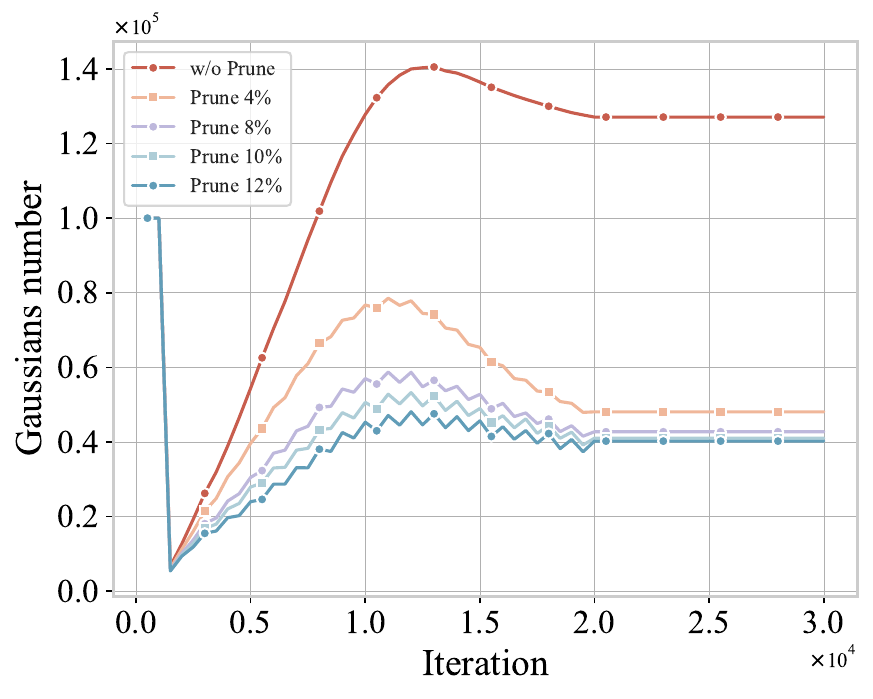}}\\
\vspace{-10pt}
\hspace{0.05cm}
\subfigure{\includegraphics[width=3.1in]{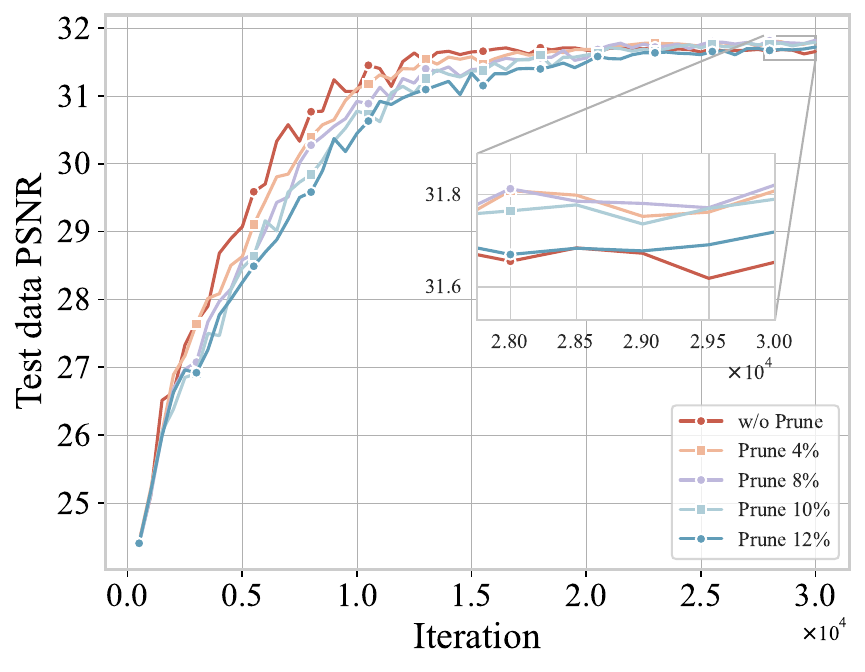}}
\vspace{-10pt}
\caption{The impact of randomly pruning Gaussians at different proportions on the number of Gaussians and the quality of generated images (30 training views).}
\vspace{-20pt}
\label{fig6}
\end{figure}

\begin{figure}[!t]
\centering
\hspace{0.2cm}\includegraphics[width=3.4in]{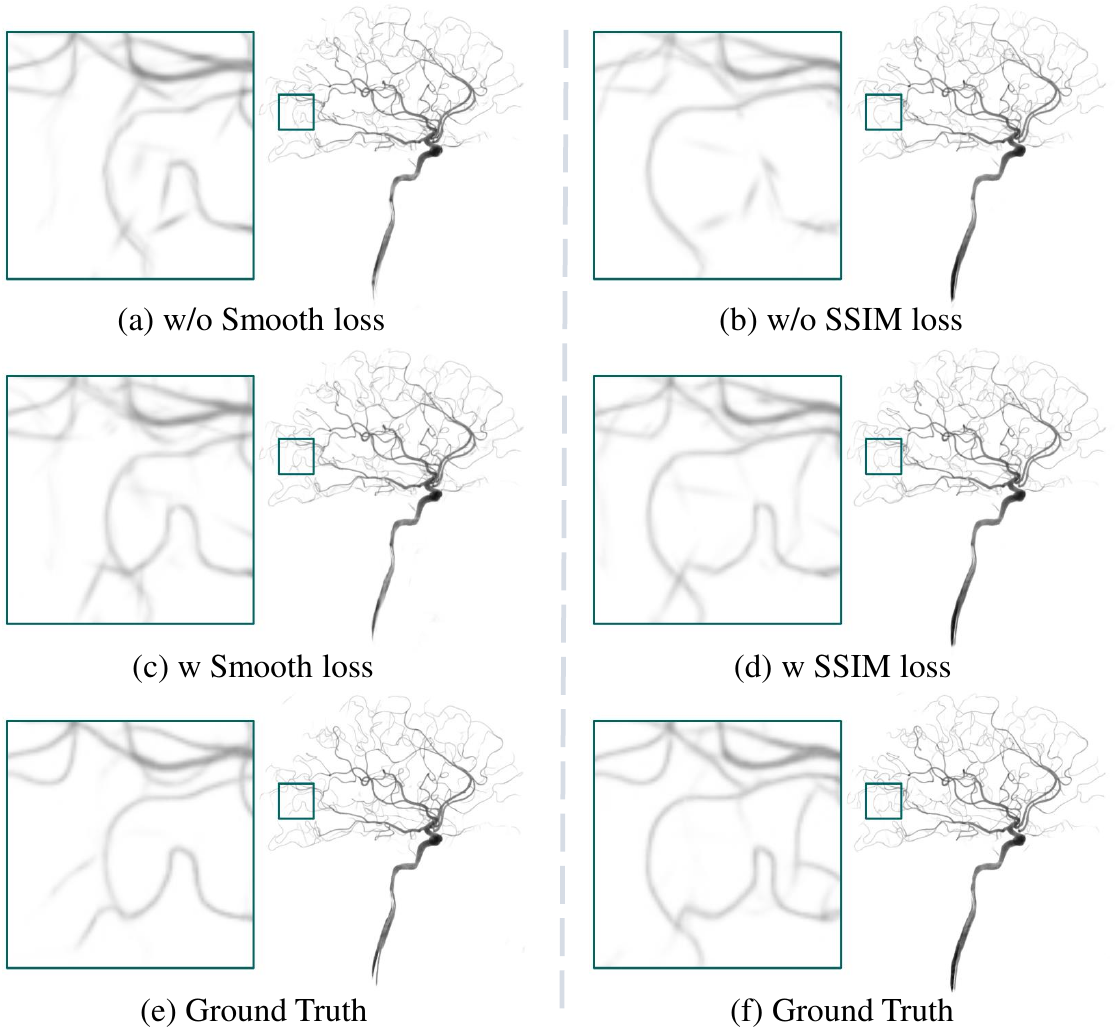}
%\vspace{-5pt}
\caption{{Visualization results of the ablation experiment of the loss function (30 training views).}}
\label{fig7}
\vspace{-10pt}
\end{figure}

\subsubsection{Ablation study for Loss function}
In this ablation experiment, we validated the effectiveness of each loss term. We compared the impact of including the Smooth loss versus excluding it on the results. The results are shown in Table~\ref{table4} and Fig.~\ref{fig7}. It can be observed that when there is a significant number of viewpoints, the improvement from incorporating the Smooth loss term is not substantial. However, when the number of viewpoints is limited, such as 10 or 20, the inclusion of the Smooth loss term effectively enhances the reconstruction metrics. This suggests that the Smooth loss term can effectively mitigate overfitting issues that arise when the number of views is restricted. From Fig.~\ref{fig7}, it is also evident that the introduction of the Smooth loss in the model is beneficial for fitting the scene. Furthermore, we removed the SSIM loss term and noticed a decrease in the evaluation metrics. From Fig.~\ref{fig7}, it can be observed that without the inclusion of SSIM loss, the model fails to capture the finer details of blood vessels accurately.

\subsubsection{Ablation study for pruning strategies}
We conducted ablation experiments on different pruning strategies, where the Gaussians were pruned based on their initial opacity, average opacity, or maximum opacity. Specifically, when the opacity of a Gaussian drops below a certain threshold, it is pruned. The experimental results are shown in Table~\ref{table5}. From the table, it is evident that directly pruning Gaussians based on their initial opacity leads to subpar reconstruction performance. However, when using the maximum opacity and average opacity of Gaussians with an appropriate threshold, both methods yield higher reconstruction quality. This further highlights the advantages of our approach over the deformable field-based method. With our method, we can easily obtain the opacity of Gaussians at a specific time, allowing us to control the density of Gaussians accordingly.

\subsubsection{Ablation study for Temporal resolutions}
We also investigated the influence of the length of the opacity offset table on the reconstruction results, and the experimental results are shown in Tab.~\ref{tab6}. From the table, it can be observed that the quality of the rendered images gradually increases with the increase in temporal resolution. However, the gains from increasing temporal resolution are limited. For example, when the number of viewpoints is low (10), once the temporal resolution reaches a certain value (7), further increases do not lead to improvements in the metrics. We also report the number of Gaussians at different temporal resolutions in Tab.~\ref{tab6}. As shown in the table, the number of Gaussians gradually decreases with the increase in temporal resolution. This is because the expressive capability of a single Gaussian increases with higher temporal resolution, leading to a reduced number of Gaussians needed to represent the scene.

\subsection{Discussion}
\subsubsection{Analysis of the Method's Robustness}
\begin{figure*}[!h]
\centering
\includegraphics[width=6.7in]{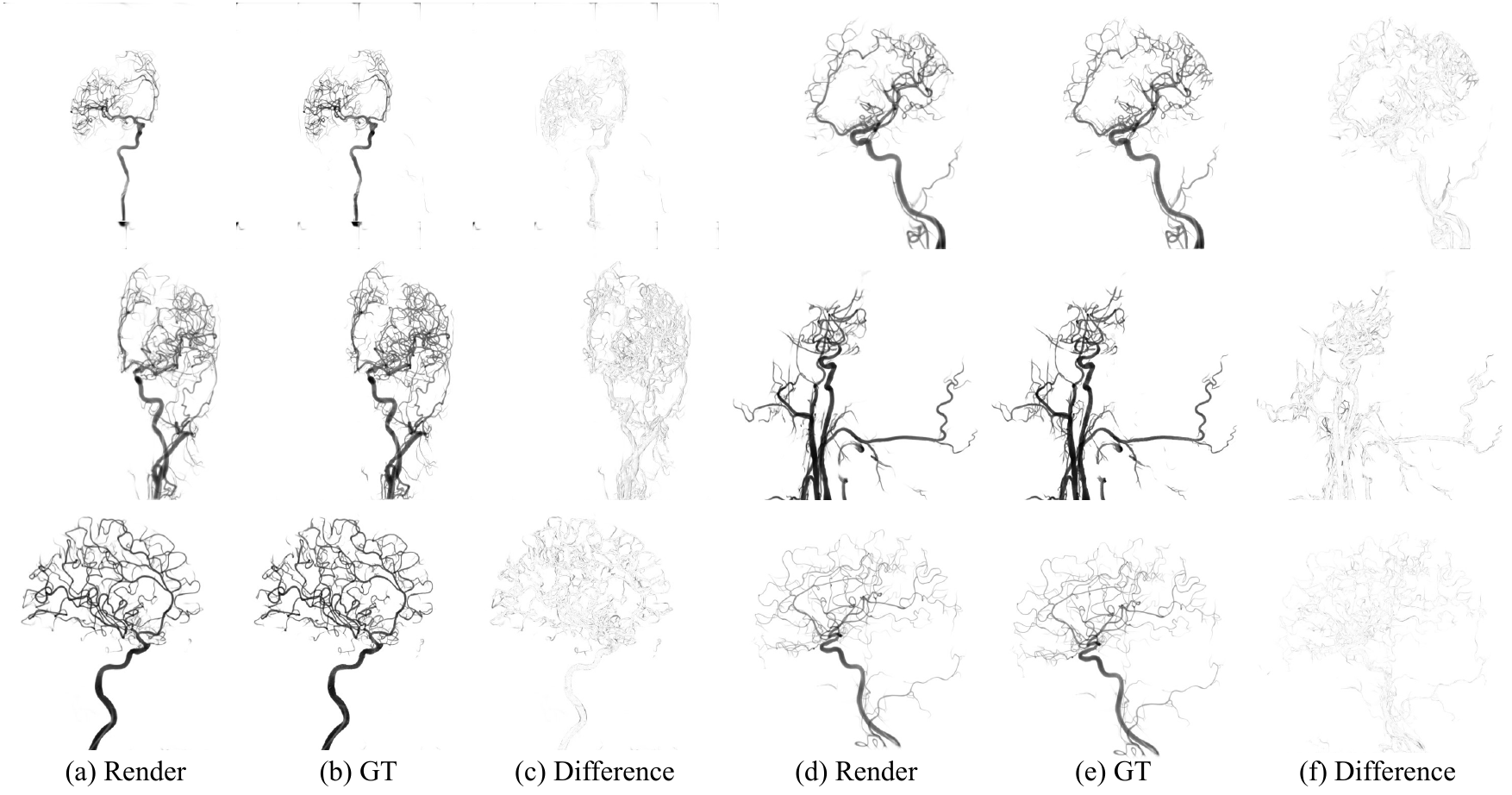}
\vspace{-5pt}
\caption{{Visualization results of experiments with data from different devices. (a) and (d) represent the novel view images rendered by the model, (b) and (e) represent the ground truth, and (c) and (f) show the difference images obtained by subtracting the rendered images from the ground truth images.}}
\label{fig:diff}
\vspace{-10pt}
\end{figure*}

To verify the robustness of the proposed model, in addition to the data from 8 patients used for the comparative experiments, we obtained additional data from 12 patients at the Union Hospital, Tongji Medical College, Huazhong University of Science and Technology. This data comes from different devices and parameter settings. We conducted experiments on this data, and the results are reported in Tab~\ref{tab:data_sources}. When the sparse ratio is 1:4, the PSNR of the results is basically above 30.  In Fig.~\ref{fig:diff}, we show the experimental results on data from these different sources. As seen in Fig.~\ref{fig:diff}, the difference between the rendered novel view images and the ground truth is very small. Our experimental results demonstrate that our method exhibits promising performance, good generalization, and robustness across these datasets.

\begin{table*}[htbp]
    \centering
    \caption{{Experimental results from different devices (Sparse ratio: Number of training views/ Number of test views).}}
    \label{tab:data_sources}
    \begin{tabular}{cccccccccccc}
        \toprule
        \ \multirow{2}{*}{Data Source }  &\multicolumn{3}{c}{Sparse ratio: 1/3 } &\multicolumn{3}{c}{Sparse ratio: 1/4 }&\multicolumn{3}{c}{Sparse ratio: 1/7 }\\& PSNR & SSIM & LPIPS &PSNR & SSIM & LPIPS &PSNR & SSIM & LPIPS\\
        \midrule
        1   &31.92 & 0.975 & 0.061& 31.66 &0.975 &0.061& 30.83 &0.974 &0.067\\
        2     &31.86 & 0.971 & 0.077& 31.44 & 0.971 & 0.075&30.94 & 0.968 & 0.081\\   
        3     &30.17 & 0.972 & 0.043    &30.06 & 0.971& 0.044 &28.88 & 0.964 & 0.052\\
        4     &29.54 & 0.957 & 0.083 &29.08 & 0.954 & 0.085 &28.58 & 0.950 & 0.088\\
        5     &33.35 & 0.973 & 0.044 &32.71 & 0.969 & 0.048 &31.96 & 0.964 & 0.049\\
        6    &30.23 & 0.958 & 0.058&30.10 & 0.957 & 0.060&28.75 & 0.945 & 0.067\\
        7    &30.50 & 0.953 & 0.069 &30.18 & 0.950 & 0.070 &29.00 & 0.938 & 0.079\\
        8    &32.71 & 0.965 & 0.053 &32.67 & 0.964 & 0.052 &31.61 & 0.955 & 0.058\\
        9   &32.63 & 0.975 & 0.057 &32.62 & 0.975 & 0.060 &31.86 & 0.972 & 0.059\\
        10    &31.30 & 0.966 & 0.060 &31.08 & 0.964 & 0.063 &30.32 & 0.959 & 0.063\\
        11     &32.72 & 0.965 & 0.053 &32.64 & 0.964 & 0.053 &31.71 & 0.956& 0.057\\
        12    &30.38 & 0.971& 0.057 &30.27 & 0.969& 0.060 &29.60 & 0.965& 0.060\\

        \bottomrule
    \end{tabular}
    \vspace{-5pt}
\end{table*}

\subsubsection{Experiment on un-denoised data}
\begin{figure}[!t]
\centering
\includegraphics[width=3.4in]{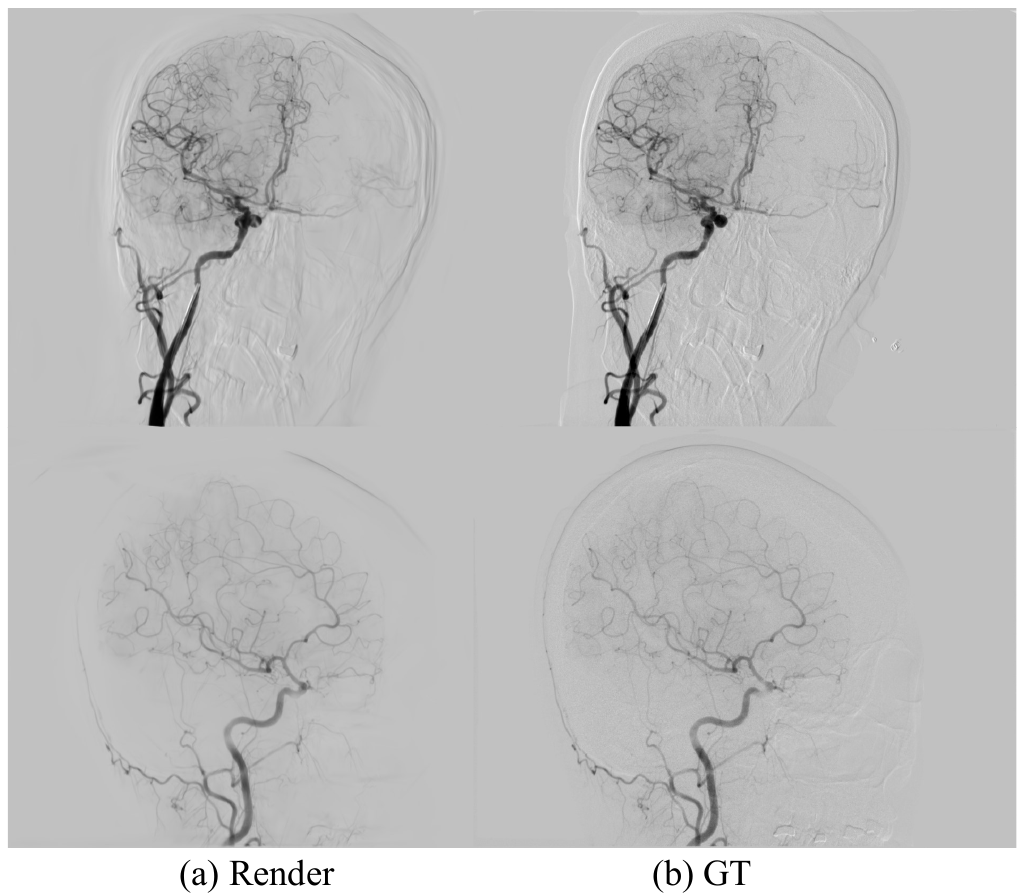}
%\vspace{-5pt}
\caption{{Visualization experimental results of the data without denoising (30 train views).}}
\label{fig:wosegment}
\vspace{-5pt}
\end{figure}

To verify whether the model can be directly applied to process un-denoised original images, we conducted experiments on the original images without denoising. As shown in Fig.\ref{fig:wosegment}, the blood vessel areas can still be rendered effectively. This indicates that when the bone signal noise is not too strong, the proposed model can be directly applied to process un-denoised images.

\section{Clinical application analysis}

\subsection{Potential clinical application value}
The proposed method has great clinical value. a) Significantly reducing patient radiation exposure: By synthesizing novel views from fewer inputs, TOGS can greatly reduce the number of DSA imaging viewpoints required, thereby substantially lowering the radiation dose experienced by patients. This is especially important for patients who require multiple diagnostic examinations. b) Enhancing the observation and analysis of cerebrovascular diseases: TOGS can synthesize more viewpoint images, providing doctors with more comprehensive brain vascular imaging data, which aids in more accurate observation, diagnosis, and analysis of cerebrovascular diseases. c) Real-time rendering capability enables interactive visualization: TOGS achieves a rendering speed of over 100 FPS, supporting interactive visualization and analysis of 4D DSA data, which can provide better diagnostic and treatment decision support for doctors.

\vspace{-5pt}
\subsection{Clinical workflow}
We can easily integrate TOGS into practical clinical workflows. An additional deep-learning workstation with more than 6GB of video memory can be equipped, and the TOGS method can be easily deployed on the workstation. After the patients' raw DICOM data is collected by the DSA imaging device, the raw data can be imported into the workstation, and the image information and corresponding pose information can be extracted using the Python scripts provided in our project, and then fed into the TOGS training and rendering process. At the same time, TOGS has low storage overhead. The Gaussian file generated by the model for each patient is typically less than 20 MB. TOGS has the capability of real-time rendering, and there are already many visualization software based on Gaussian Splatting methods available. Further development of TOGS-specific visualization software can be carried out to support the visualization of DSA results.

\vspace{-5pt}
\section{Limitations and Future Work}

Although the proposed method achieves good results under the 4D DSA new view synthesis task, there are still some limitations that need to be addressed in future work. 3DGS is an explicit discrete representation method with strong fitting capabilities.  The 3DGS-based methods are prone to overfitting issues in sparse-view scenarios. 4D DSA is a dynamic scene, and compared to static 3D scenes, the overfitting problem becomes more severe under sparse training views. In this paper, we solve this problem by introducing a Smooth loss term in the loss function and randomly pruning some Gaussians. However, in extremely sparse views, such as only four training views, the model still suffers from severe overfitting issues. In 4D DSA, due to the limited information provided by sparse views and the strong fitting capabilities of 3DGS, the model can quickly achieve a good fit with the trained views during training. Consequently, it struggles to continue training effectively, resulting in subpar performance from other views. To address this issue, it may be beneficial to explore methods for reducing the degrees of freedom in dynamic Gaussian Splatting approaches, such as the Smooth Loss proposed in this paper, or to investigate more efficient representations.

Rendering images at different time frames from the same viewpoint remains a challenging problem. Due to the limitation of having only one image per time frame for a given viewpoint in the dataset, there is a lack of supervision from other time frames. As a result, rendering images from other time frames can suffer from issues such as blurring and ghosting. This is essentially an overfitting problem in the temporal dimension caused by data sparsity. In the future, there is a possibility of incorporating depth prior information~\cite{zhu2023fsgs}, adding extra supervision~\cite{lao2023corresnerf}, and exploring other approaches to address this issue.

\iffalse
\begin{figure}[!t]
\centering
\hspace{0.42cm}\includegraphics[width=3.4in]{fig7.pdf}
%\vspace{-20pt}
\caption{The rendering results of different time frames from the same viewpoint.}
\label{fig7}
\vspace{-10pt}
\end{figure}
\fi

In addition to addressing the aforementioned challenges, future work should further explore efficient 3D/4D representation methods specifically tailored to the unique field of medical imaging. This paper takes a straightforward approach using Gaussians with explicit opacity offset tables to model 4D DSA data. In the future, it is crucial to explore more efficient methods for modeling the intensity variations of contrast agents. Additionally, the current Gaussian Splatting method still faces challenges such as overfitting and difficulties in surface extraction~\cite{guedon2023sugar}. Future research should focus on refining the Gaussian Splatting method or exploring more efficient representation and rendering techniques.

%\vspace{-5pt}
\section{Conclusion}
In this paper, to achieve high-quality and real-time rendering in 4D DSA novel view synthesis under sparse views, we propose TOGS. The variation in radiation signal intensity of contrast agents in 4D DSA leads to changes in the opacity of the scene. We introduce opacity offset tables in 3DGS to model the opacity variation in the scene. By querying the opacity offset table, we obtain the opacity of the Gaussian at a given moment. Afterward, the image at that moment can be rendered using splatting. We introduced a Smoothing loss and implemented random pruning to control the density of Gaussians. By implementing these two strategies, we successfully alleviated the issue of model overfitting and improved the quality of rendered images. The proposed method effectively enhances the quality of novel view synthesis in 4D DSA. Moreover, it has the advantages of real-time rendering and low memory overhead. In the future, we will further explore more efficient representations and rendering methods for 4D DSA novel view synthesis.

\vspace{-5pt}
\section*{References}

\vspace{-10pt}

\bibliographystyle{ieeetr}
\bibliography{ref}

\end{document}